%% file: main.tex
\documentclass[10pt,twocolumn,letterpaper]{article}

\usepackage[pagenumbers]{cvpr} 


\definecolor{cvprblue}{rgb}{0.21,0.49,0.74}
\usepackage[pagebackref,breaklinks,colorlinks,allcolors=cvprblue]{hyperref}
\usepackage{colortbl}
\usepackage{multirow}
\usepackage{multicol}

\def\paperID{17584} 
\def\confName{CVPR}
\def\confYear{2025}

\title{MapGS: Generalizable Pretraining and Data Augmentation for Online Mapping via Novel View Synthesis}

\author{
  Hengyuan Zhang\textsuperscript{1} \quad David Paz\textsuperscript{2} \quad Yuliang Guo\textsuperscript{2} \quad Xinyu Huang\textsuperscript{2} \\ Henrik I. Christensen\textsuperscript{1} \quad Liu Ren\textsuperscript{2}\\
  \textsuperscript{1}Contextual Robotics Institute, UC San Diego \\
  \textsuperscript{2}Bosch North America \\
  {\tt\small \{hyzhang, hichristensen\}@ucsd.edu }\\ {\tt\small \{david.pazruiz, yuliang.guo2, xinyu.huang, liu.ren\}@us.bosch.com }
}

\begin{document}
\twocolumn[{%
\renewcommand\twocolumn[1][]{#1}%
\maketitle
\begin{center}
    \centering
     \includegraphics[width=1.0\textwidth]{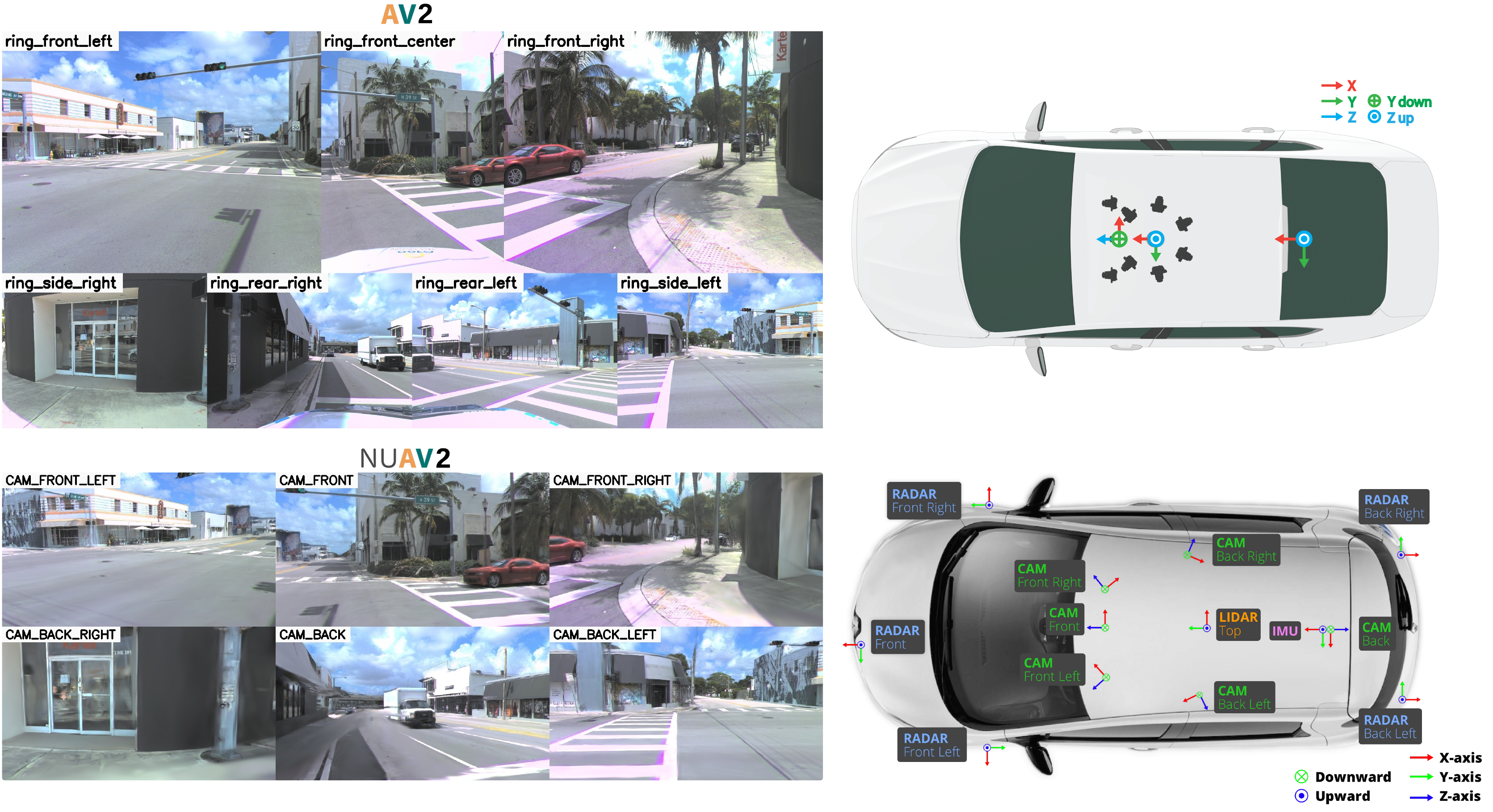}
     \captionsetup{type=figure}
     \captionof{figure}{\textbf{Cross sensor data alignment.} Online mapping algorithms struggle when deployed on a vehicle with different sensor configurations and require labeled data with the same sensor configuration. With the source sensor configuration images collected by Argoverse 2 (AV2)~\cite{Argoverse2} data collection vehicles (top row), we propose to leverage Gaussian splatting to render images in the target nuScenes (NUSC)~\cite{caesar_nuscenes_2020} sensor configuration (bottom row). The synthesized dataset, named nuAV2, is used to train online mapping algorithms to reduce the generalization gap using different training paradigms.}
     \label{fig:mapgs-teaser}
\end{center}%
}]
\input{sec/0_abstract}    
\input{sec/1_intro}

\input{sec/2_relatedwork}
\input{sec/3_method}
\input{sec/4_experiment}
\input{sec/5_conclusion}
{
    \small
    \bibliographystyle{ieeenat_fullname}
    \bibliography{main}
}

\input{sec/X_suppl}

\end{document}

%% file: sec/0_abstract.tex
\begin{abstract}
Online mapping reduces the reliance of autonomous vehicles on high-definition (HD) maps, significantly enhancing scalability. However, recent advancements often overlook cross-sensor configuration generalization, leading to performance degradation when models are deployed on vehicles with different camera intrinsics and extrinsics. With the rapid evolution of novel view synthesis methods, we investigate the extent to which these techniques can be leveraged to address the sensor configuration generalization challenge. We propose a novel framework leveraging Gaussian splatting to reconstruct scenes and render camera images in target sensor configurations. The target config sensor data, along with labels mapped to the target config, are used to train online mapping models. Our proposed framework on the nuScenes and Argoverse 2 datasets demonstrates a performance improvement of 18\% through effective dataset augmentation, achieves faster convergence and efficient training, and exceeds state-of-the-art performance when using only 25\% of the original training data. This enables data reuse and reduces the need for laborious data labeling. Project page at \url{https://henryzhangzhy.github.io/mapgs}.
\end{abstract}

%% file: sec/1_intro.tex
\section{Introduction}
\label{sec:intro}

Novel developments for multi-modality sensor fusion have enabled new perception methods for autonomous driving applications in the past few years. The developments include topics on 3D object detection~\cite{liu2024ray,jiang2022polar,bevstereo}, online map generation and reasoning~\cite{li2023toponet,luo2024smerf,wu2024topomlp}, occupancy prediction~\cite{sima2023_occnet,viewformer}, and in the form of end-to-end frameworks~\cite{jiang2023vad,hu2023_uniad}. State-of-the-art results have been achieved across these tasks and in real-world settings; however, significant challenges remain in achieving sensor and platform generalization.

For example, a perception model trained on the NuScenes (NUSC)~\cite{caesar_nuscenes_2020} dataset does not readily generalize to arbitrary sensor and platform configurations such as the Waymo Open~\cite{Sun2020waymo} or the Argoverse 2 (AV2)~\cite{Argoverse2} configurations. The sensor configurations may differ in the number of cameras and LiDARs, their location on the car, orientation, and intrinsic properties such as camera focal length and field of view. Difference in sensor configurations has been shown to lead to performance drop across 3D perception tasks, such as detection~\cite{zheng2023cross}, segmentation~\cite{philion2020lift, wang2023towards} and mapping~\cite{Ranganatha2024SemVecNet}. Though there remain other domain gap issues to additionally explore, achieving generalization across sensor configuration remains a significant challenge that also limits the amount of data that can be used for development. In the worst of cases, separate datasets have to be curated and labeled to retrain or fine tune deep learning architectures when the sensor setups change. This leads to lengthy development processes to deploy models across vehicle types such as sedans, SUVs, and trucks.

Recent works have begun to address the challenge of sensor generalization from two primary approaches: model-based and data-driven solutions. From a model perspective, methods aim to incorporate sensor-agnostic properties~\cite{chang2024udga, li2024unidrive} either in problem formulation or directly within the architecture~\cite{Ranganatha2024SemVecNet}. In contrast, the data-driven approach focuses on aligning camera parameters or applying data regeneration techniques~\cite{zheng2023cross, tzofi2023viewpoint}. While promising results have been achieved, developing general strategies for surround-view online HD map generation~\cite{MapTR, maptrv2} with real data remains an open problem. 

To this end, we propose a data-based paradigm using Gaussian splatting~\cite{kerbl3Dgaussians, yan2024streetgs} for scene reconstruction and novel view synthesis, aimed at reducing sensor configuration gaps in 3D online mapping algorithms. This approach includes a recipe for data regeneration given new custom configurations, where Gaussian splatting is applied to reconstruct static and dynamic scenes. The approach proposed is extended to the AV2 dataset, generating new camera data in the NUSC sensor configuration. The results show that the approach enhances generalization, improves data and training efficiency, and enables data reusability for surround-view online mapping. In summary, our key contributions are as follows.

\begin{enumerate}
    \item We propose a paradigm of using Gaussian splatting to reconstruct scenes and render novel views to reduce sensor configuration gaps for 3D online mapping algorithms.
    \item The recipe for novel view synthesis and data regeneration is used to construct a new dataset termed nuAV2 data; where we reconstruct scenes in the AV2 dataset and regenerate camera data in the NUSC sensor configuration. We will make the code and data public.
    \item Three key performance and efficiency benefits arise from nuAV2: i) effective dataset augmentation to improve generalization, where a performance improvement of 18\% is observed, ii) more efficient training and faster convergence, iii) reduction of training data requirements, where state-of-the-art performance is exceeded with only 25\% of the original target sensor configuration training data.

\end{enumerate}

%% file: sec/2_relatedwork.tex
\section{Related Work}

\textbf{3D Reconstruction.} Going beyond sparse reconstruction~\cite{Hartley_Zisserman_2004_multiple, Agarwal2009rome}, recent advancements in neural rendering has significantly improved dense 3D reconstruction and photo-realistic rendering~\cite{mildenhall2020nerf, kerbl3Dgaussians}. Notably, Kerbl \etal~\cite{kerbl3Dgaussians} propose using 3D Gaussians to represent the scene and a fast splatting rendering algorithm that enables fast training and real-time photo-realistic rendering. Initially designed for static and bounded scenes, this approach has been extended to dynamic and open environments. For autonomous driving scenes, Chen \etal~\cite{chen2023periodic} propose to add periodic vibrations into the Gaussian center and opacity. This approach increases the model parameters, leading to high-quality rendering for dynamic scenes but overfitting to viewpoints close to the trajectory. Yan \etal~\cite{yan2024streetgs} in Street Gaussian leverages tracks from the labeled dataset to separate dynamic objects and static backgrounds. Dynamic objects are treated as static objects moving along the tracks. The separation reduces the problem to a static reconstruction problem, generating more robust novel views. \\

\noindent\textbf{Perception Models and Generalization.} Recent techniques show promising performance and have achieved state-of-the-art generalization performance on various perception tasks. For instance, UDGA~\cite{chang2024udga} introduces a multi-view overlap depth constraint and employs a label-efficient domain adaptation component. The approach requires a few labels and fine-tuning. A separate approach termed Lift-Splat-Shoot (LSS)~\cite{philion2020lift} approaches BEV from an arbitrary number of cameras by individually lifting each image into frustums and projecting them into BEV, which enhances robustness to calibration errors. On the other hand, BEVFormer~\cite{li2022bevformer} leverages camera parameters to sample 3D points, which are then used to cross-attend perspective view (PV) and BEV features. Though the approach does not generalize across different sensor configurations, it achieves robust performance within the operation domain. Additionally, Wang \etal~\cite{wang2023towards} analyzes the causes of domain gaps in 3D detection tasks, decouples depth estimation from camera intrinsic parameters, and introduces an adversarial training loss to improve domain generalization. \\

\noindent\textbf{Data Approach to Generalization.} A separate line of work focuses on sensor generalization improvements by aligning data representations. For instance, SemVectNet~\cite{Ranganatha2024SemVecNet} introduces an intermediate rasterized semantic map representation to bridge the gap between sensor configurations, using this semantic map to regress vectorized lane elements. UniDrive~\cite{li2024unidrive} employs unified virtual cameras and proposes a ground-aware projection method to address challenges related to parameter variability, with validation conducted in CARLA simulation. The GGS method~\cite{han2024ggs} introduces Gaussian Splatting techniques for virtual lane change generation using a single-view camera, applying a diffusion loss to enhance the approach. On the other hand, mesh reconstruction techniques based on monocular depth estimation have been introduced to reduce the gap in BEV segmentation tasks~\cite{tzofi2023viewpoint}. Lastly, camera normalization methods have also been incorporated to improve robustness in detection tasks. For instance, \cite{zheng2023cross} assesses the generalization of camera-based 3D object detectors, showing that simply aggregating datasets leads to performance drops, indicating that adding more data alone is insufficient. This approach introduces sensor alignment methods, including camera intrinsics synchronization, camera extrinsics correction, and ego frame alignment.

In contrast with previous methods, our work seeks to provide a new perspective to the following questions: i) can we leverage novel view synthesis strategies to bridge the gap between unseen viewpoints from different sensor configurations? ii) if a gap persists, can we mitigate the generalization challenges specific to multi-view cameras? iii) can we formulate a data regeneration recipe to reuse existing datasets and perform more sophisticated data augmentations?

%% file: sec/3_method.tex
\section{Method}

\begin{figure*}[t]
    \centering
    \includegraphics[width=0.98\linewidth]{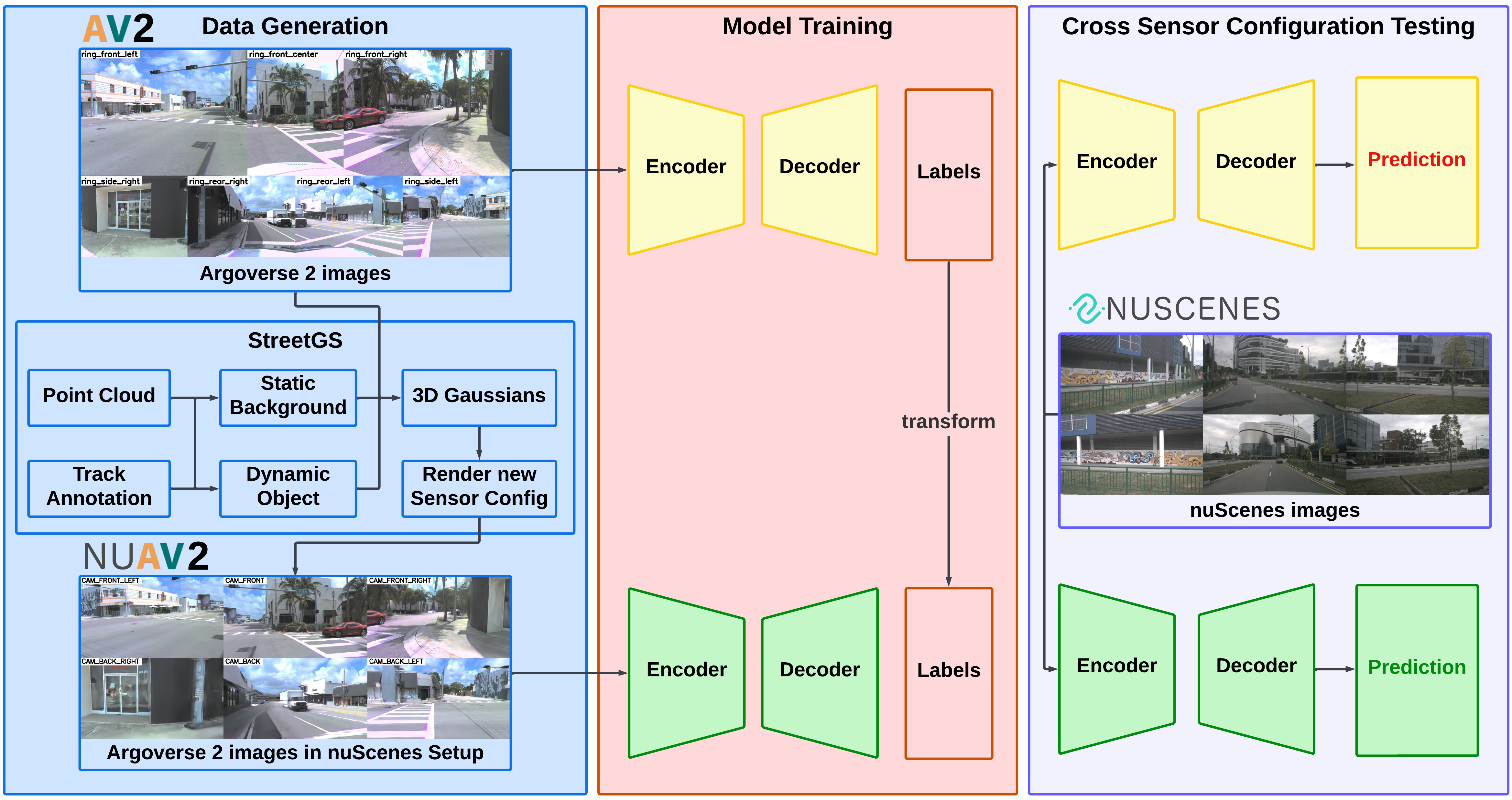}

    \caption{\textbf{MapGS Pipeline.} Deploying online mapping models on a different sensor configuration is challenging. MapGS proposes to leverage Street Gaussian (StreetGS) to reconstruct the scene, then render into images in target sensor configuration. We then train a model with this data and labels. Finally, we test the model in target sensor configuration.}
    \label{fig:mapgs-pipeline}
\end{figure*}

Generalizing online mapping algorithms to different sensor configurations is challenging. We propose a new paradigm to generate data in different sensor configurations using Gaussian splatting to reduce the generalization gap. As shown in \cref{fig:mapgs-pipeline}, for an autonomous driving dataset, we first reconstruct the scene with existing sensor data using Gaussian splatting, then render images and map labels for a target sensor configuration, and finally use this data as part of a training protocol. The following subsections explain each step in detail.

We focus on a camera-only online mapping problem, which aims to find a model $\mathcal{M}$ that maps images $\mathbf{I}=\mathbf{I}_1, \dots, \mathbf{I}_{N_c}$ into vector map elements $\mathbf{M}=\mathbf{M}_1, \dots, \mathbf{M}_{N_m}$. $N_c$ denotes the number of images from the surround-view setup in autonomous driving. $N_m$ denotes the number of map elements. Each map element $\mathbf{M}_i$ consists of $N_p$ coordinate points $(x_1, y_1), \dots, (x_{N_p}, y_{N_p})$ and a class label $c_i$, such as $divider$, $crossing$, $boundary$ and $centerline$.

\subsection{Scene Reconstruction}

We first process the sensor data's source images $\mathbf{I}^S$ and reconstruct a Scene $\mathcal{G}$. The original 3D Gaussian splatting~\cite{kerbl3Dgaussians} requires many images to reconstruct a scene. Reconstructing autonomous driving scenes with dynamic objects, open environments and homogeneous views remains quite challenging. We base our solution on the Street Gaussian (StreetGS)~\cite{yan2024streetgs} approach, which leverages prior knowledge from labeled tracks and point clouds to reconstruct outdoor scenes. The prior constraints the optimization process and produce high-quality novel views.

StreetGS separates each scene into static background $\mathcal{B}$ and dynamic vehicles $\mathcal{D}$ based on annotated tracks. It preprocesses the LiDAR points clouds $\mathbf{P}_t$ to filter out points inside the bounding boxes. The remaining points from all frames are aggregated based on pose to initialize a static background $\mathcal{B}$. The dynamic vehicles $\mathcal{D}$ are treated as static vehicles moving along the tracks. Hence the point cloud within each annotated bounding box is cropped and aggregated into a single box for each track. This point cloud initializes a 3D Gaussian for the object. A sky embedding $\mathcal{S}$ is also defined to capture the sky and far away objects.

During the optimization process, the object 3D Gaussians are transformed to corresponding bounding boxes according to the tracks. Given a training camera view, the static background, object Gaussians and sky are rasterized to render an image. The image is then compared to the ground truth image and optimize the 3D Gaussians and the sky. Ego vehicle masks are added to the ground truth images that capture the ego vehicle to prevent generating floaters.

This approach leverages prior to optimize a dynamic scene similarly to when it is static, without introducing additional temporal parameters. As we show in \cref{sec:exp-data-gen}, this reduces the distortion from overfitting when the viewpoint changes drastically.

\subsection{Target Configuration Data Generation}

With the reconstructed scene $\mathcal{G}$, we render images in the target configuration, denoted as $\mathbf{I}^T$. The process tries to render images as if we are driving the target vehicle in the scene. 

Given the source vehicle pose ${}^{\mathcal{G}}\mathbf{T}_{V_S}$ at time $t$, target camera to vehicle pose ${}^{V_T}\mathbf{T}_{C_T}$, and target vehicle to source vehicle transform ${}^{V_S}\mathbf{T}_{V_T}$. The target vehicle pose is 
$${}^{\mathcal{G}}\mathbf{T}_{V_T} = {}^{\mathcal{G}}\mathbf{T}_{V_S} {}^{V_S}\mathbf{T}_{V_T}.$$ Note that this transform is necessary as vehicle frames are not consistent across datasets. For example, the NUSC dataset~\cite{caesar_nuscenes_2020} defines the vehicle frame on the rear axle projected on the ground whereas the AV2~\cite{Argoverse2} dataset defines the vehicle frame on the rear axle around $33cm$ above the ground.

Then the target camera pose ${}^{\mathcal{G}}\mathbf{T}_{C_T}$ is defined by $${}^{\mathcal{G}}\mathbf{T}_{C_T} = {}^{\mathcal{G}}\mathbf{T}_{V_T} {}^{V_T}\mathbf{T}_{C_T}.$$ Along with the target camera intrinsic $\mathbf{K}_T$, we can render images $\mathbf{I}$ in the target sensor configurations. 

Map labels are also transformed into the new sensor setup for ego-centric mapping training and evaluation.

\subsection{Online Mapping}

We only focus on the map element geometry without considering the recently popular topology reasoning~\cite{wang2023openlanev2}. We leverage a camera-only version of MapTRv2~\cite{maptrv2}. MapTRv2 consists of three major components, a feature extractor, a view-transform module and a vector map decoder. The feature extractor takes surround view images $\mathbf{I}$ as input, extracts perspective features $\mathbf{F}_{PV}$. Then the view transform module projects them into BEV features $\mathbf{F}_{BEV}$. Finally, the Deformable DETR~\cite{zhu2021deformable} decoder decodes vector map elements from the BEV features.

%% file: sec/4_experiment.tex
\section{Experiment}
\label{sec:exp}

\begin{table*}[!t]\centering
    \footnotesize
    \begin{tabular}{cccccccc}
        
        \toprule
        
        \multirow{2}{*}{Experiment} & \multirow{2}{*}{Modality} & \multirow{2}{*}{Training Config} & \multicolumn{4}{c}{Class} & \cellcolor{gray!30}\\
        
        &&&Div. &Cross. &Bound. &Center. &\cellcolor{gray!30} \multirow{-2}{*}{mAP}\\
        
        \cmidrule{1-8}
        
        \multirow{3}{*}{Direct Generalization} &Camera+LiDAR &AV2 (SemVecNet~\cite{Ranganatha2024SemVecNet}) &8.7 &5.0 &15.7 &19.4 &\cellcolor{gray!30} 12.2\\
        
        \cmidrule{2-8}
        
        &Camera & AV2 4 epochs &1.3 &0.4 &1.0 &2.1 &\cellcolor{gray!30} 1.2\\
        
        &Camera &nuAV2 4 epochs &\textbf{3.1} &\textbf{3.2} &\textbf{6.4} &\textbf{12.4} &\cellcolor{gray!30} \textbf{6.3}\\
        
        \cmidrule{1-8}
        
        \multirow{2}{*}{Fine-tuning} &Camera & AV2 4 epochs + NUSC 32 epochs &12.2 &4.6 &22.1 &24.2 &\cellcolor{gray!30} 15.8 \\ 
        
        &Camera &nuAV2 4 epochs + NUSC 32 epochs &\underline{\textbf{21.2}} &\textbf{18.9} &\textbf{31.6} &\textbf{28.7} &\cellcolor{gray!30} \textbf{25.1} \\ 
        
        \cmidrule{1-8}
        
        Joint training &Camera & nuAV2 + NUSC merged 32 epochs &19.4 &\underline{20.5} &\underline{32.5} &\underline{29.8} &\cellcolor{gray!30} \underline{25.5} \\

        \cmidrule{1-8}
        
        Oracle &Camera & NUSC 32 epoch &16.5 &12.7 &30.6 &26.5 &\cellcolor{gray!30} 21.6 \\
        
        \bottomrule
        
    \end{tabular}
    \caption{\textbf{Online mapping performance on the NUSC eval dataset.} MapTRv2 trained on nuAV2 generated by MapGS sets a new baseline for camera-only online mapping generalization. The nuAV2 datasets can also augment small datasets with pretraining and joint training. \textbf{bold} indicates higher performance in comparison group and \underline{underline} indicates the overall highest performance.
    }\label{tab:mapgs}
    \vspace{-6pt}
\end{table*}

\subsection{Data Generation}
\label{sec:exp-data-gen}

We perform experiments on the AV2~\cite{Argoverse2} and NUSC~\cite{caesar_nuscenes_2020} datasets. The AV2 dataset is chosen as the source sensor configuration and NUSC as the target configuration. AV2 is equipped with 7 ring cameras positioned near the LiDAR sensor, whereas the 6 NUSC cameras are mounted along the roof edges. Additionally, the cameras on each platform vary in viewing direction and intrinsic parameters as shown in \cref{fig:mapgs-teaser}--the camera configurations differ significantly between the two data collection vehicles. 

\begin{figure}[t]
    \centering
    \includegraphics[width=0.98\linewidth]{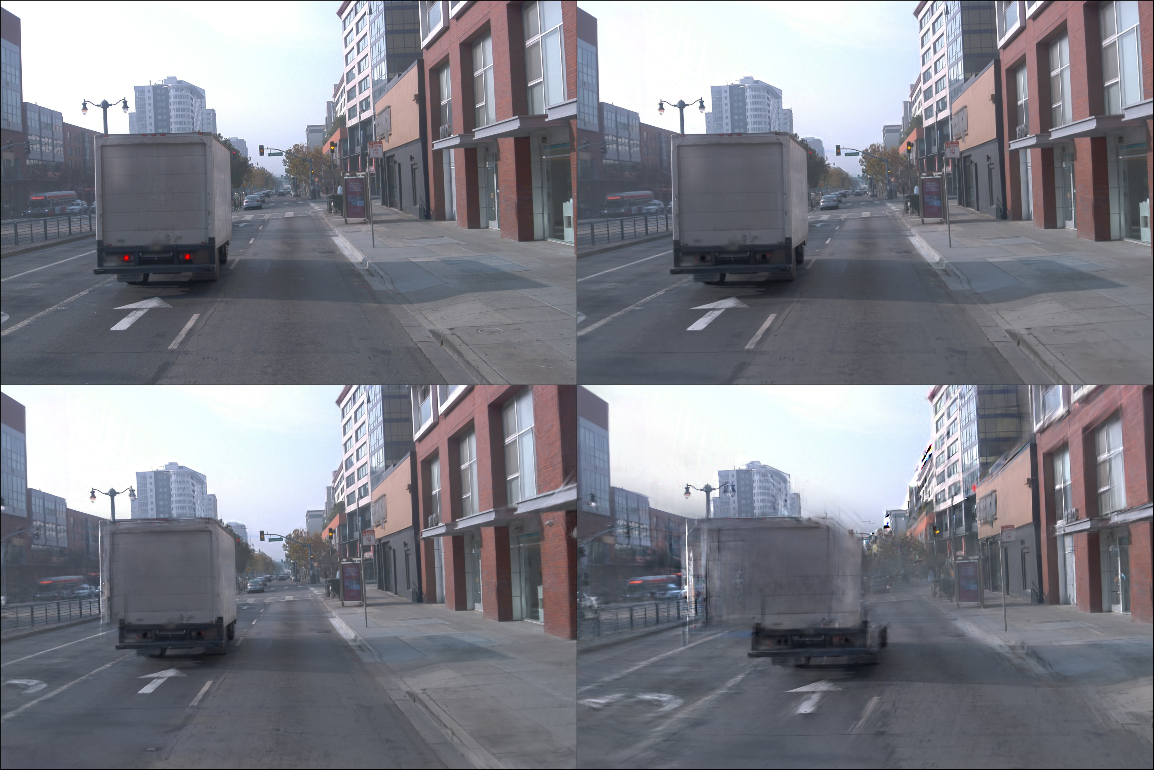}

    \caption{\textbf{PVG Distortion.} While the reconstruction along the trajectory (top right) has high quality compared to groudtruth (top left), moving the camera away from the trajectory, such as 1 m backwards (bottom left) and 1 m left (bottom right) causes the quality to drop significantly.}
    \label{fig:pvg}
\end{figure}

 To reconstruct scenes from AV2 images and render images in the target NUSC sensor configuration we experiment with two Gaussian Splatting techniques. The first 3D reconstruction technique is based on Periodic Vibration Gaussian (PVG)~\cite{chen2023periodic}; PVG introduces periodic vibration-based temporal dynamics to the 3D Gaussian splatting and reconstructs high-quality images along the trajectory. However, significant reconstruction degradation is observed when the cameras deviate from the original trajectory. As shown in \cref{fig:pvg}, the reconstruction quality is high when the camera is placed at a frame excluded in training but when the camera is shifted 1 meter backward, the results start to deteriorate. Furthermore, vehicles become distorted when we shift the camera to its left, largely deviating from its trajectory. We hypothesize that this is due to the high number of parameters in PVG that leads to overfitting to specific views.

\begin{figure}[t]
    \centering
    \includegraphics[width=0.98\linewidth]{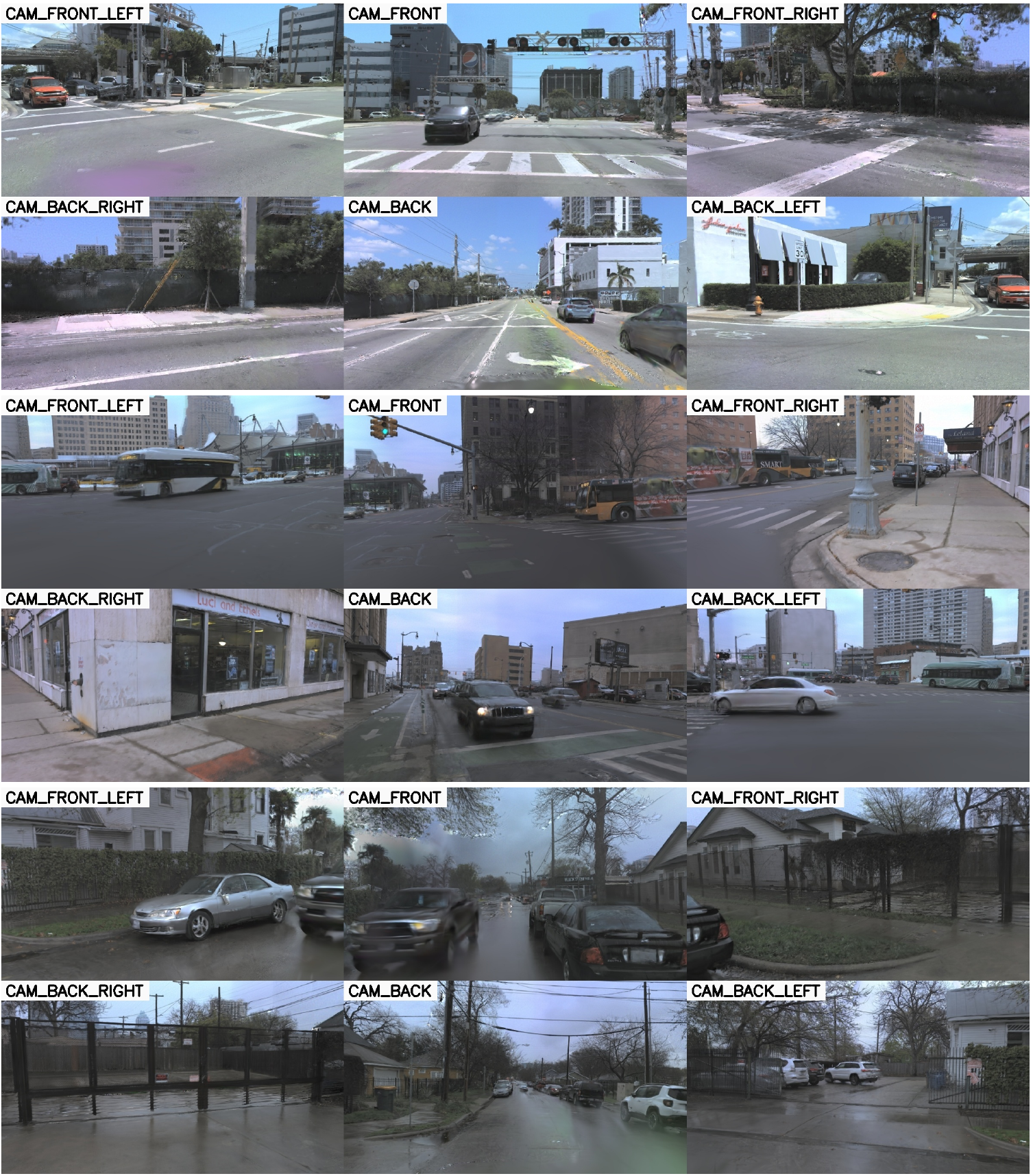}

    \caption{\textbf{nuAV2 Examples.} nuAV2 renders reconstructed AV2 dataset in the NUSC sensor configuration.}
    \label{fig:mapgs_examples}
\end{figure}

In contrast, Street Gaussian (StreetGS)~\cite{yan2024streetgs} leverages 3D bounding box labels from the dynamic scene to construct a static background and static vehicle Gaussian moving with its track labels. We show rendered images from the nuAV2 dataset along with its original AV2 data in \cref{fig:mapgs-teaser} and more novel view examples in \cref{fig:mapgs_examples}. Given the quality of reconstructions and advantages over PVG, we base our solution on the StreetGS approach.

While the number of cameras, placement and angles changes, the rendering generates coherent surround-view images as if the scene is traversed by a NUSC data collection vehicle. Notably, since the AV2 front center camera is rotated, the novel front view contains information from all three front cameras, and the quality remains. This is also true for the novel rear view that fuses the original rear left and rear right images. 

We adapted the StreetGS implementation from LightwheelAI~\cite{LightwheelAI} to the AV2 dataset, reconstructed the 700 training and 150 validation scenes. Only images associated with synchronized LiDAR frames are used, resulting in approximately 156 frames per scene. Additionally, we corrected the sky rotating in the wrong direction issue and added ego vehicle masks to improve the quality and consistency. Then we render the images in NUSC sensor configuration, resulting in 655k images for the training set and 140k images for the validation set. Only the training set is used for later experiments and all evaluations are performed on the NUSC validation set. In the cases where we use the NUSC dataset, we follow the geo-disjoint splits defined in~\cite{lilja2024localizationevaluatedataleakage} to reduce data leakage in evaluation.

To simplify the training and evaluation process, we converted the data into NUSC dataset format, named nuAV2. nuAV2 allows the data to be treated as a version of the NUSC dataset and used with existing code easily. More importantly, we can train the model in nuAV2 and evaluate it on the NUSC validation set without any conversions.

\subsection{Metrics} The vector maps are evaluated based on mean average precision (mAP) under Chamfer Distance following~\cite{MapTR, maptrv2}. 

The Chamfer distance describes the distance between a predicted map element ${\mathbf{x}_1, \dots, \mathbf{x}_n}$ and a ground truth map element ${\mathbf{y}_1, \dots, \mathbf{y}_m}$ without considering their order, given by $$D_{\text{Chamfer}} = \frac{1}{2n} \sum_{i=1}^{n} \min_{j} || \mathbf{x}_i - \mathbf{y}_j || + \frac{1}{2m} \sum_{j=1}^{m} \min_{i} || \mathbf{y}_j - \mathbf{x}_i ||.$$
In our setup, $m=n$ as the deformable DETR decoder~\cite{zhu2021deformable} is configured to predict the same number of points as in the ground truth. Predictions are associated with ground truth for each class based on the Chamfer distance, and the average precision is computed at three distance thresholds: 0.5 m, 1.0 m, and 1.5 m. The final mean average precision (mAP) is calculated by averaging AP across four classes, $divider$, $crossing$, $boundary$, and $centerline$. To enhance readability, all mAP values are multiplied by 100.

\subsection{Direct Generalization}
\label{sec:direct-generalization}
While large autonomous driving datasets are now available, online mapping models generalize poorly when deployed on vehicles with different camera setups. Thus, we first aim to answer: to what extent can novel-view synthesis methods reduce the sensor gap?

As an attempt to answer this question, we first train a MapTRv2 model on the nuAV2 dataset and test its direct generalization performance on NUSC evaluation data. We additionally compare a MapTRv2 model trained on AV2 and also tested on NUSC. As shown in \cref{tab:mapgs}, the nuAV2 model achieves 6 mAP, which is a $4.25\times$ improvement. Indicating that sensor alignment through data regeneration provides clear generalization benefits. An additional comparison that we make is with respect to a MapTRv2 model trained and evaluated on the NUSC train and evaluation sets, respectively; the model is termed Oracle. In comparison, our nuAV2 model achieves 29\% of the performance compared to the Oracle model. A significant factor for this gap pertains to potential domain shifts within the training distributions. As a result, sensor-specific generalization cannot be directly quantified. Though the comparison results are not direct, we present them for reference.

Another approach compared is SemVecNet~\cite{Ranganatha2024SemVecNet}; which achieves 12.2 mAP using intermediate semantic map rasters based on a camera and LiDAR setup. Though the approach presents generalization benefits, our approach sets a new baseline for a camera-only surround-view cross-sensor setup. 

\subsection{Dataset Alignment}

\begin{table}[!t]\centering
    \footnotesize
    \begin{tabular}{ccccccc}
    \toprule
    NUSC & nuAV2 & \multicolumn{4}{c}{Class} & \cellcolor{gray!30} \\
    Percent & Pretrain & Div. & Cross. & Bound. & Center. & \cellcolor{gray!30} \multirow{-2}{*}{mAP} \\
    \cmidrule{1-7}
    \multirow{2}{*}{5} & no & 9.9 & 8.1 & 20.8 & 19.6 & \cellcolor{gray!30} 14.6 \\
    & yes & \textbf{15.6} & \textbf{12.9} & \textbf{25.4} & \textbf{26.4} & \cellcolor{gray!30} \textbf{20.1} \\
    \cmidrule{1-7}
    \multirow{2}{*}{10} & no & 12.5 & 9.4 & 24.2 & 23.3 & \cellcolor{gray!30} 17.3 \\
    & yes & \textbf{15.3} & \textbf{16.4} & \textbf{26.9} & \textbf{25.8} & \cellcolor{gray!30} \textbf{21.1} \\
    \cmidrule{1-7}
    \multirow{2}{*}{25} & no & 14.4 & 13.6 & 27.7 & 26.0 & \cellcolor{gray!30} 20.4 \\
    & yes & \textbf{18.4} & \textbf{16.8} & \textbf{29.4} & \textbf{27.5} & \cellcolor{gray!30} \textbf{23.0} \\
    \cmidrule{1-7}
    \multirow{2}{*}{100} & no & 16.5 & 12.7 & 30.6 & 26.5 & \cellcolor{gray!30} 21.6 \\
    & yes & \textbf{19.8} & \textbf{18.1} & \textbf{32.2} & \textbf{28.6} & \cellcolor{gray!30} \textbf{24.7} \\
    \bottomrule
    \end{tabular}
    \caption{\textbf{Enhancing small datasets with nuAV2.} Pretraining on nuAV2 consistently improves the model performance. 
    }\label{tab:enhance-low-data}
    \vspace{-6pt}
\end{table}

Besides cases without data or labels, it is common to label a small subset of the data for finetuning. We then ask whether our approach can align the datasets to transfer knowledge and reduce groundtruth label requirements. More specifically, a model pretrained on our nuAV2 dataset and fine-tuned on a smaller subset can maintain the performance or even improve the model performance.

First, we randomly sample three subsets of the geo-disjoint NUSC training set, comprising 5\%, 10\%, and 25\% of the data. We then pretrain a MapTRv2 model on nuAV2 for 8 epochs, and fine-tune using the different splits including 100\% NUSC dataset. We also train baseline models from scratch with the sampled subsets and the full NUSC dataset. The model trained from scratch with 100\% of the data is defined as the Oracle. As the dataset size decreases with sampling, additional epochs are required to reach the same number of iterations. To account for this, we define an \textit{equivalent epoch} for each subset, representing the number of epochs needed to match the iteration count of a single epoch on the full NUSC dataset. Accordingly, one \textit{equivalent epoch} on the 5\%, 10\%, and 25\% subsets corresponds to 20, 10, and 4 actual epochs, respectively.

As shown in \cref{tab:enhance-low-data}, nuAV2 improves the model performance when only a small subset of labeled data is available. In general, pretraining on nuAV2 brings 2-3 mAP improvements across different experiments. In fact, with nuAV2, a model fine-tuned on only 25\% of the NUSC dataset outperforms the Oracle model training on the full NUSC dataset. The final performance of a model fine-tuned on 100\% of the NUSC dataset surpasses the Oracle model performance, showing its ability to transfer knowledge from other datasets collected by different sensor setups. Even with only 5\% data, nuAV2 pretraining brings the performance from 67.5\% to 93.1\% of the Oracle model performance.

\begin{figure}[t]
    \centering
    \includegraphics[width=0.98\linewidth]{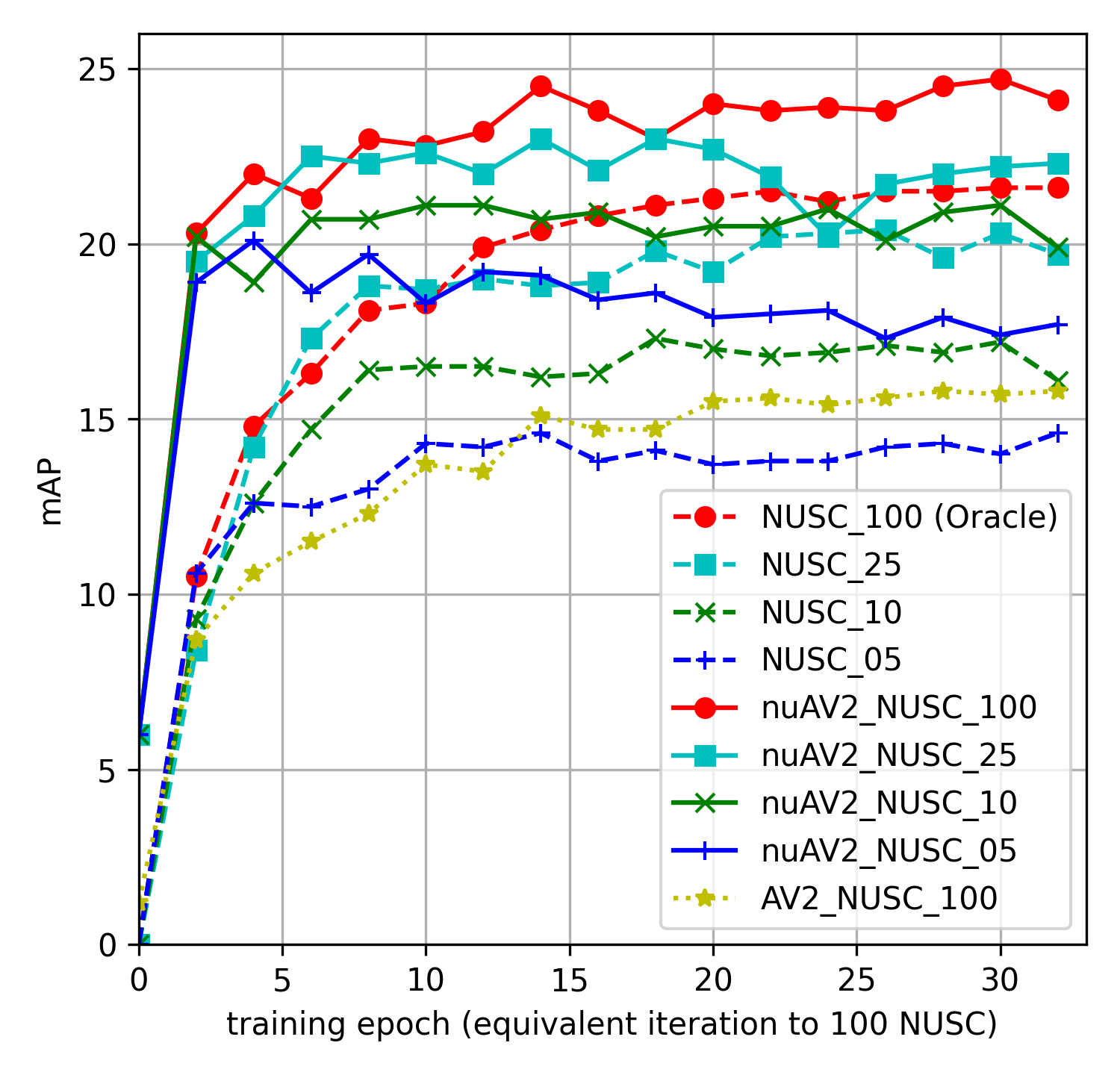}

    \caption{\textbf{Low data fine-tuning.} Models pretrained with nuAV2 achieve high performance rapidly, often surpassing their baseline convergence performance, whereas pretraining with AV2 not only slows down training but also reduces overall performance.
    }
    \label{fig:less-data}
\end{figure}

We additionally pretrain a version on the AV2 dataset, then fine-tune it on NUSC. In contrast, we find that training convergence significantly slows down compared to the Oracle model and reduces overall performance, as shown in \cref{fig:less-data}. This made it evident that pretraining on nuAV2 with the same sensor configuration is key to the performance gain.

\begin{figure}[t]
    \centering
    \includegraphics[width=0.98\linewidth]{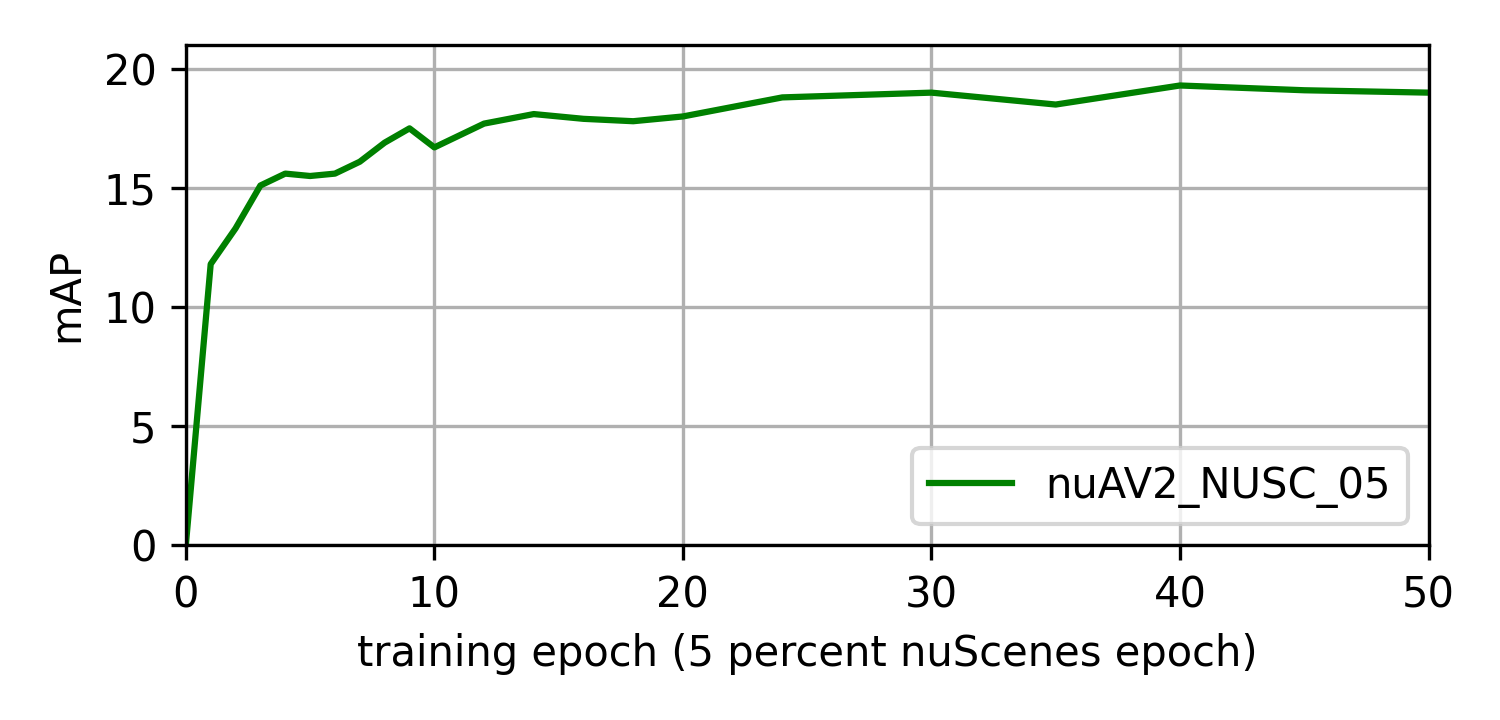}

    \caption{\textbf{5\% NUSC fine-tuning.} Models reach over 75\% of the final performance with 3 epochs fine-tuning on 5\% of the NUSC dataset, showing the effectiveness of pretraining on the nuAV2 dataset.}
    \label{fig:fast-convergence}
\end{figure}

Interestingly, all fine-tuning experiments, regardless of the dataset size, reach higher than 18 mAP performance within 2 equivalent epochs, as shown in \cref{fig:less-data}. For the 5\% data fine-tuning experiments, the performance at epoch 3 is higher than the convergence performance only trained on 5\% NUSC, as shown in \cref{fig:fast-convergence}. This suggests that a small amount of data was able to align the model with the existing setup and unleash the power of the pretrained model. 

We also find that the performance for fine-tuning on 5\% NUSC started to decline rather than increase with more training on the subset in the target domain. This suggests that the model might overfit to this small subset while starting with nuAV2 boosts and improves generalization.

\begin{figure}[t]
    \centering
    \includegraphics[width=0.98\linewidth]{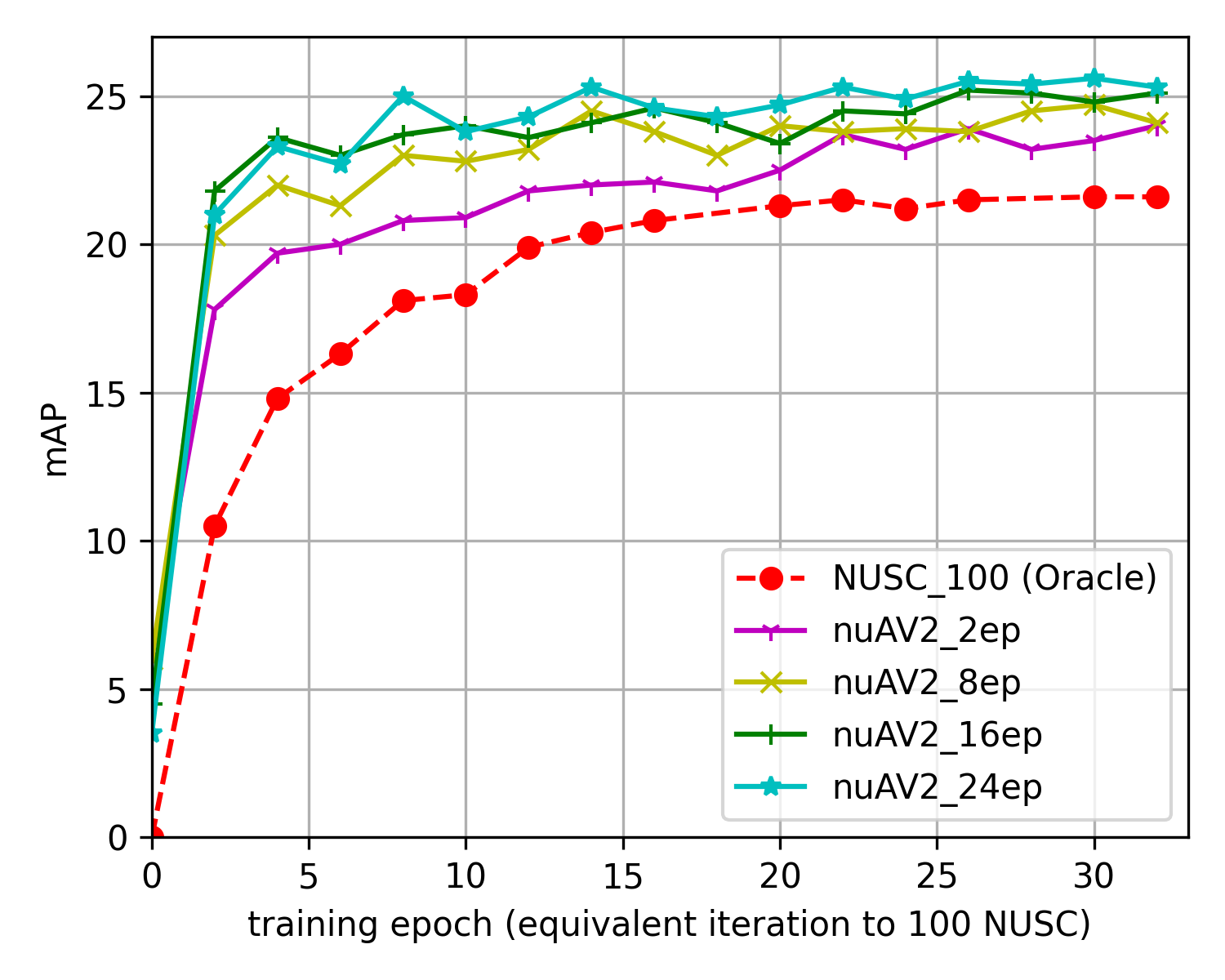}

    \caption{\textbf{Fine-tuning performance across different pretraining epochs.} Pretraining on the nuAV2 dataset enhances final performance, with additional epochs accelerating convergence.
    }
    \label{fig:prior_epoch}
\end{figure}

\noindent\textbf{Ablation on a pretrained model at a specific epoch.} Pretraining with just 2 epochs can improve the model performance significantly, though training for more epochs leads to slightly higher performance, as shown in \cref{fig:prior_epoch}. It is worth noting that this is true even when the pretraining starts to overfit on the nuAV2 dataset and direct generalization performance drops, as shown in the second column of \cref{tab:prior-influence}. 

\begin{table}[!t]\centering
    \footnotesize
    \begin{tabular}{ccccccc}
        
        \toprule
        
        nuAV2 & Direct & \multicolumn{4}{c}{Class} &\cellcolor{gray!30} \\
        
        Epoch & mAP &Div. &Cross. &Bound. &Center. & \cellcolor{gray!30} \multirow{-2}{*}{mAP}\\
        
        \cmidrule{1-7}
        2 &\textbf{6.1} &19.5 &18.5 &31.0 &26.9 &\cellcolor{gray!30} 24.0\\

        8 &6.0 &19.8 &18.1 &32.2 &28.6 &\cellcolor{gray!30} 24.7\\
        
        16 &4.5 &19.7 &19.4 &\textbf{32.6} &29.1 &\cellcolor{gray!30} 25.2\\

        24 &3.5 &\textbf{21.0} &\textbf{19.7} &32.4 &\textbf{29.5} &\cellcolor{gray!30} \textbf{25.6}\\
        
        \bottomrule
        
    \end{tabular}
    \caption{\textbf{Impact of additional pretraining epochs on nuAV2.} While increasing pretraining epochs leads to overfitting on nuAV2, causing a decline in direct generalization mAP (Direct mAP), the fine-tuning performance increases.
    }\label{tab:prior-influence}
    \vspace{-6pt}
\end{table}

\subsection{Joint Training}

With the nuAV2 dataset, we can also easily mix it with the NUSC dataset and perform joint training. In this way, we can treat the MapGS as an augmentation technique and nuAV2 becomes the augmented data.

Since an epoch in the merged dataset has similar iterations compared to 5 epochs in NUSC, we noticed that the network converges slower. This is reasonable as the dataset size is larger and signal from NUSC is mixed with signals from AV2. But the model performance increases to 25.5 mAP, surpassing the oracle performance trained only on the NUSC dataset by 18\%.

\subsection{Limitations}

While our approach sets a new baseline in camera-only sensor configuration generalization, there are some limitations we would like to note. 

\begin{figure}[t]
    \centering
    \includegraphics[width=0.98\linewidth]{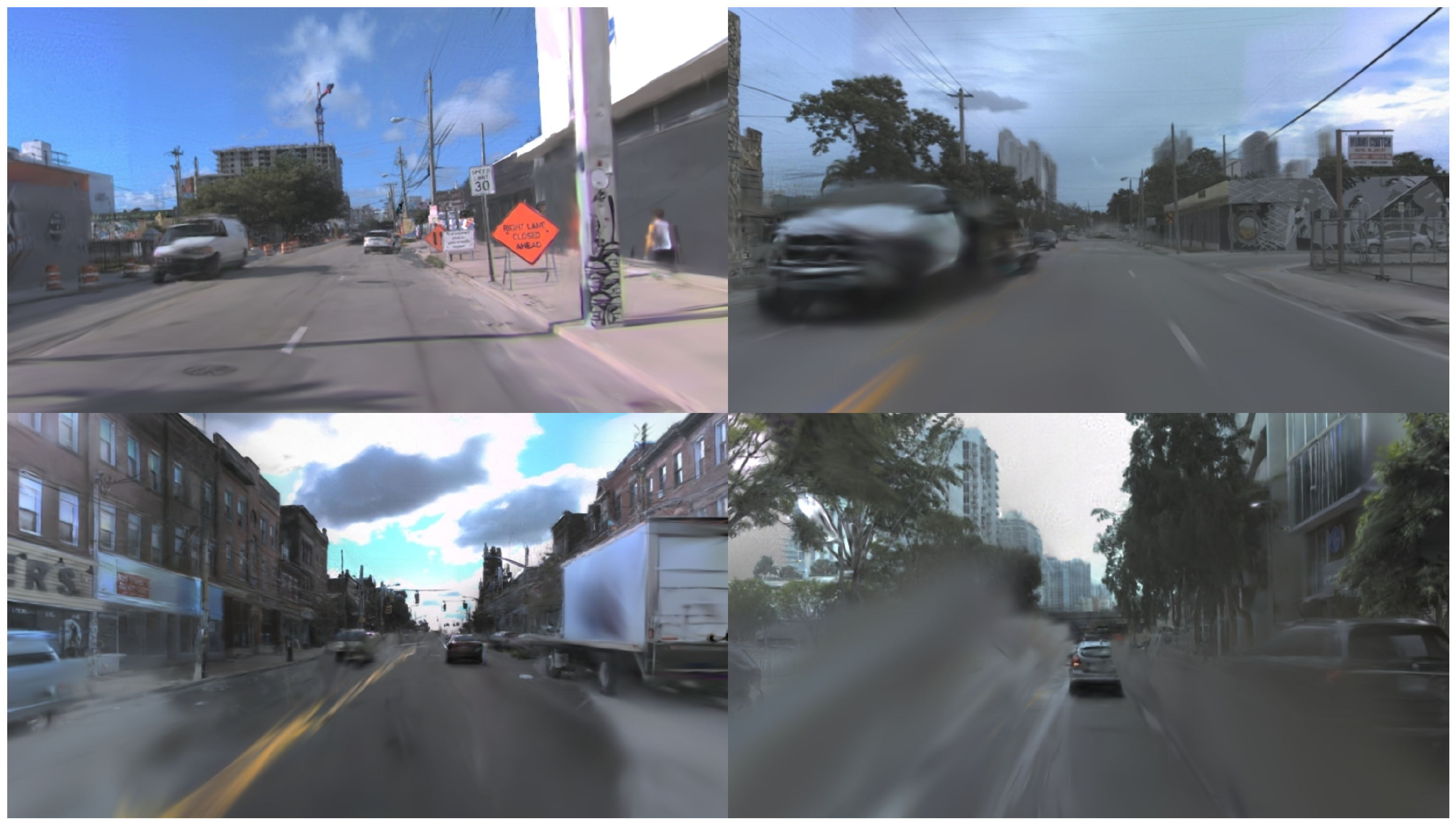}

    \caption{\textbf{MapGS Challenges.}. Challenges exist in reconstructing dynamic objects such as pedestrians (top left), and vehicles (top right). Lighting changes and camera auto gain can lead to uneven lighting (bottom left) and floaters (bottom right).}
    \label{fig:mapgs_challenges}
\end{figure}

First, the approach is bottlenecked by the reconstruction quality. The render quality will drop significantly if the render sensor configuration deviates so much from the original view that some content is not visible in the original data. For example, raising the camera height to the height of the truck will allow much visibility through occlusion but this may not be recovered from data collected by a sedan. Additionally, deformable objects are not handled thus pedestrians will not render well. For outdoor scenes, changes in lighting and the auto gain from cameras can potentially lead to floaters in the reconstructed Scenes, as shown in \cref{fig:mapgs_challenges}

Second, our evaluation doesn't separate sensor configuration generalization from data domain generalization. As outlined in Sec.~\ref{sec:direct-generalization}, the gap from the AV2 dataset to the NUSC dataset is more than just sensors, but also the different road markings, road layout, and urban environments.

%% file: sec/5_conclusion.tex
\section{Conclusion}

We propose MapGS, a framework to address the lack of data in training online mapping models for specific sensor configurations. MapGS leverages Gaussian splatting to reconstruct scenes and render images in target sensor configuration views. From this work, we propose a recipe for data regeneration and introduce the nuAV2 dataset, converting images from the Argoverse 2 dataset into the nuScenes dataset views for training. We show that the approach reduces the gap between sensor configurations. Moreover, it is an effective dataset augmentation technique and enables pretraining or joint training in online mapping tasks.

%% file: sec/X_suppl.tex
\clearpage
\setcounter{page}{1}
\maketitlesupplementary

\section{Qualitative results}

\begin{figure}[t]
    \centering
    \includegraphics[width=0.98\linewidth]{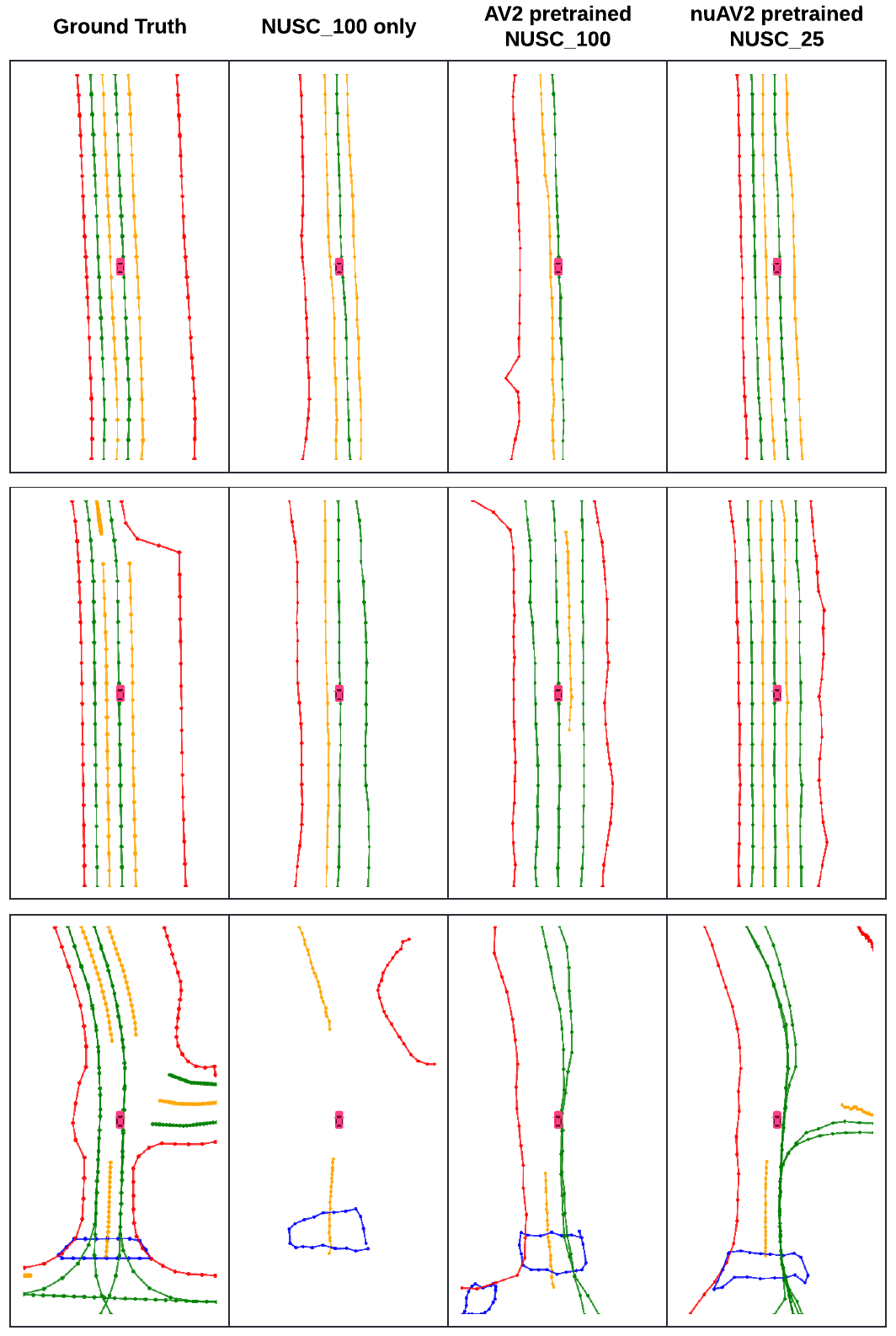}

    \caption{\textbf{Qualitative results of MapGS.} Map elements are color-coded as follows: $divider$ (yellow), $boundary$ (red), $crossing$ (blue), and $centerline$ (green). NUSC\_100 and NUSC\_25 represent 100\% and 25\% of the NUSC dataset, respectively. The first two rows highlight the performance gains achieved through pretraining with nuAV2. The last row illustrates online mapping remains challenging, particularly when using a small backbone to test on geo-disjoint splits.}
    \label{fig:mapgs_mapping}
\end{figure}

Qualitative results for online mapping are presented in \cref{fig:mapgs_mapping}. MapTRv2~\cite{maptrv2}, pretrained on the nuAV2 dataset and fine-tuned with 25\% nuScenes (NUSC)~\cite{caesar_nuscenes_2020} dataset, surpasses the Oracle model, trained exclusively on 100\% NUSC data. In contrast, pretraining on the Argoverse 2 (AV2)~\cite{Argoverse2} dataset leads to a decline in performance.

It is worth noting that online mapping remains challenging, especially with a small backbone and testing on geo-disjoint splits~\cite{lilja2024localizationevaluatedataleakage}. In many scenarios, the network struggles to capture the complexity of the scene.

\section{Additional details of experiments}

\subsection{nuAV2 in NUSC format}

To facilitate compatibility with the NUSC development tools, we package the nuAV2 dataset into the NUSC database format. 

In addition to adopting the NUSC camera setup, we align the vehicle and LiDAR frames in nuAV2 to match the NUSC coordinate system. Although LiDAR data is not directly used for the camera-only online mapping task, the LiDAR frame is used in MapTRv2 as a reference for the map center. To ensure alignment, we create a virtual LiDAR in nuAV2, derived from the \texttt{LIDAR\_TOP} configuration in NUSC. This virtual LiDAR is used solely to align the map centers between the datasets.

These modifications allow seamless pretraining on nuAV2 and fine-tuning on NUSC without requiring any additional adjustments. Furthermore, they enable joint training across both datasets, streamlining the integration process.

\subsection{AV2 pretraining}

Although the overall architecture of MapTRv2 is designed to handle sensor configuration changes, some minor adjustments are required to train a single model sequentially on AV2 and NUSC. First, we remove the camera embeddings from MapTRv2. These embeddings, learned for each specific camera, are tied to a fixed number of cameras and therefore not transferable across datasets with differing sensor setups. Second, we eliminate the perspective-view segmentation loss, as it also relies on a fixed camera configuration.

Additionally, similar to nuAV2, we create a virtual LiDAR frame to align the map center for pretraining on AV2. 

With these modifications, a model pretrained on AV2 can seamlessly be used for evaluation or fine-tuning on the NUSC dataset.

\subsection{Computation and storage}

The primary computational demands of our pipeline occur during preprocessing, reconstruction, and training.\\

\noindent\textbf{Preprocessing.} Preprocessing involves generating semantic masks for each image, extracting sky and object masks, aggregating point clouds for the static background, and assigning points to objects based on track annotations. By initializing 3D Gaussians with LiDAR point clouds instead of those derived from structure-from-motion, we reduce the preprocessing time from 5–6 hours to approximately 1 hour per segment on an NVIDIA A10 GPU. For 700 training and 150 validation segments, this step requires around 5 days on an 8-GPU node.\\

\noindent\textbf{Reconstruction.} Reconstructing the scenes is the most computationally intensive step. An AV2 scene with 7 cameras and 156 frames takes approximately 6 hours. Using four 8-GPU nodes, the reconstruction process for all scenes takes about one week.\\

\noindent\textbf{Training.} Training MapTRv2 on the nuAV2 dataset for 8 epochs requires roughly 1 day on an 8-GPU node equipped with A10 GPUs. Fine-tuning the model on the NUSC dataset for 32 epochs also takes approximately 1 day.\\

While the reconstruction step is computationally expensive, it offers the advantage of reusability. Once reconstructed, the scenes can be adapted for new sensor configurations with minimal effort. For example, creating a dataset for pretraining with the Waymo Open Dataset~\cite{Sun2020waymo} would only require updating camera positions and rendering novel views.\\

\noindent\textbf{Storage requirements.} The size of reconstructed 3D Gaussian splatting scenes varies between 700 MB and 3.5 GB, depending on travel distance and scene complexity, with a total dataset size of approximately 1.7 TB. Rendered images for new sensor configurations require an additional 180 GB of disk storage.

%% file: main.bbl
\begin{thebibliography}{34}
\providecommand{\natexlab}[1]{#1}
\providecommand{\url}[1]{\texttt{#1}}
\expandafter\ifx\csname urlstyle\endcsname\relax
  \providecommand{\doi}[1]{doi: #1}\else
  \providecommand{\doi}{doi: \begingroup \urlstyle{rm}\Url}\fi

\bibitem[Agarwal et~al.(2009)Agarwal, Snavely, Simon, Seitz, and Szeliski]{Agarwal2009rome}
Sameer Agarwal, Noah Snavely, Ian Simon, Steven~M. Seitz, and Richard Szeliski.
\newblock Building rome in a day.
\newblock In \emph{2009 IEEE 12th International Conference on Computer Vision}, pages 72--79, 2009.

\bibitem[Caesar et~al.(2020)Caesar, Bankiti, Lang, Vora, Liong, Xu, Krishnan, Pan, Baldan, and Beijbom]{caesar_nuscenes_2020}
Holger Caesar, Varun Bankiti, Alex~H. Lang, Sourabh Vora, Venice~Erin Liong, Qiang Xu, Anush Krishnan, Yu Pan, Giancarlo Baldan, and Oscar Beijbom.
\newblock {nuScenes: A Multimodal Dataset for Autonomous Driving}.
\newblock In \emph{Proceedings of the IEEE/CVF Conference on Computer Vision and Pattern Recognition (CVPR)}, pages 11618--11628, 2020.

\bibitem[Chang et~al.(2024)Chang, Lee, Kim, Kim, Lee, Ji, Jang, and Kim]{chang2024udga}
Gyusam Chang, Jiwon Lee, Donghyun Kim, Jinkyu Kim, Dongwook Lee, Daehyun Ji, Sujin Jang, and Sangpil Kim.
\newblock Unified domain generalization and adaptation for multi-view 3d object detection, 2024.

\bibitem[Chen et~al.(2023)Chen, Gu, Jiang, Zhu, and Zhang]{chen2023periodic}
Yurui Chen, Chun Gu, Junzhe Jiang, Xiatian Zhu, and Li Zhang.
\newblock Periodic vibration gaussian: Dynamic urban scene reconstruction and real-time rendering.
\newblock \emph{arXiv:2311.18561}, 2023.

\bibitem[Han et~al.(2024)Han, Zhou, Long, Wang, and Xiao]{han2024ggs}
Huasong Han, Kaixuan Zhou, Xiaoxiao Long, Yusen Wang, and Chunxia Xiao.
\newblock Ggs: Generalizable gaussian splatting for lane switching in autonomous driving.
\newblock \emph{arXiv preprint arXiv:2409.02382}, 2024.

\bibitem[Hartley and Zisserman(2004)]{Hartley_Zisserman_2004_multiple}
Richard Hartley and Andrew Zisserman.
\newblock \emph{Multiple View Geometry in Computer Vision}.
\newblock Cambridge University Press, 2 edition, 2004.

\bibitem[Hu et~al.(2023)Hu, Yang, Chen, Li, Sima, Zhu, Chai, Du, Lin, Wang, Lu, Jia, Liu, Dai, Qiao, and Li]{hu2023_uniad}
Yihan Hu, Jiazhi Yang, Li Chen, Keyu Li, Chonghao Sima, Xizhou Zhu, Siqi Chai, Senyao Du, Tianwei Lin, Wenhai Wang, Lewei Lu, Xiaosong Jia, Qiang Liu, Jifeng Dai, Yu Qiao, and Hongyang Li.
\newblock {Planning-oriented Autonomous Driving}.
\newblock In \emph{Proceedings of the IEEE/CVF Conference on Computer Vision and Pattern Recognition (CVPR)}, pages 17853--17862, 2023.

\bibitem[Jiang et~al.(2023{\natexlab{a}})Jiang, Chen, Xu, Liao, Chen, Zhou, Zhang, Liu, Huang, and Wang]{jiang2023vad}
Bo Jiang, Shaoyu Chen, Qing Xu, Bencheng Liao, Jiajie Chen, Helong Zhou, Qian Zhang, Wenyu Liu, Chang Huang, and Xinggang Wang.
\newblock Vad: Vectorized scene representation for efficient autonomous driving.
\newblock \emph{Proceedings of the IEEE/CVF International Conference on Computer Vision (ICCV)}, 2023{\natexlab{a}}.

\bibitem[Jiang et~al.(2023{\natexlab{b}})Jiang, Zhang, Miao, Zhu, Gao, Hu, and Jiang]{jiang2022polar}
Yanqin Jiang, Li Zhang, Zhenwei Miao, Xiatian Zhu, Jin Gao, Weiming Hu, and Yu-Gang Jiang.
\newblock Polarformer: Multi-camera 3d object detection with polar transformers.
\newblock In \emph{AAAI}, 2023{\natexlab{b}}.

\bibitem[Kerbl et~al.(2023)Kerbl, Kopanas, Leimk{\"u}hler, and Drettakis]{kerbl3Dgaussians}
Bernhard Kerbl, Georgios Kopanas, Thomas Leimk{\"u}hler, and George Drettakis.
\newblock 3d gaussian splatting for real-time radiance field rendering.
\newblock \emph{ACM Transactions on Graphics}, 42\penalty0 (4), 2023.

\bibitem[Klinghoffer et~al.(2023)Klinghoffer, Philion, Chen, Litany, Gojcic, Joo, Raskar, Fidler, and Alvarez]{tzofi2023viewpoint}
Tzofi Klinghoffer, Jonah Philion, Wenzheng Chen, Or Litany, Zan Gojcic, Jungseock Joo, Ramesh Raskar, Sanja Fidler, and Jose~M Alvarez.
\newblock Towards viewpoint robustness in bird's eye view segmentation.
\newblock In \emph{International Conference on Computer Vision}, 2023.

\bibitem[Li et~al.(2024{\natexlab{a}})Li, He, Zhou, Cheng, Wen, and Zhang]{viewformer}
Jinke Li, Xiao He, Chonghua Zhou, Xiaoqiang Cheng, Yang Wen, and Dan Zhang.
\newblock Viewformer: Exploring spatiotemporal modeling for multi-view 3d occupancy perception via view-guided transformers.
\newblock In \emph{Proceedings of the European Conference on Computer Vision (ECCV)}, pages 90--106, Cham, 2024{\natexlab{a}}. Springer Nature Switzerland.

\bibitem[Li et~al.(2023{\natexlab{a}})Li, Chen, Wang, Li, Yang, Geng, Jiang, Wang, Xu, Xu, Yan, Luo, and Li]{li2023toponet}
Tianyu Li, Li Chen, Huijie Wang, Yang Li, Jiazhi Yang, Xiangwei Geng, Shengyin Jiang, Yuting Wang, Hang Xu, Chunjing Xu, Junchi Yan, Ping Luo, and Hongyang Li.
\newblock {Graph-based Topology Reasoning for Driving Scenes}.
\newblock \emph{arXiv preprint arXiv:2304.05277}, pages 1--12, 2023{\natexlab{a}}.

\bibitem[Li et~al.(2023{\natexlab{b}})Li, Bao, Ge, Yang, Sun, and Li]{bevstereo}
Yinhao Li, Han Bao, Zheng Ge, Jinrong Yang, Jianjian Sun, and Zeming Li.
\newblock Bevstereo: Enhancing depth estimation in multi-view 3d object detection with temporal stereo.
\newblock \emph{Proceedings of the AAAI Conference on Artificial Intelligence}, 37\penalty0 (2):\penalty0 1486--1494, 2023{\natexlab{b}}.

\bibitem[Li et~al.(2024{\natexlab{b}})Li, Zheng, Huang, and Keutzer]{li2024unidrive}
Ye Li, Wenzhao Zheng, Xiaonan Huang, and Kurt Keutzer.
\newblock Unidrive: Towards universal driving perception across camera configurations.
\newblock \emph{arXiv preprint arXiv:2410.13864}, 2024{\natexlab{b}}.

\bibitem[Li et~al.(2022)Li, Wang, Li, Xie, Sima, Lu, Qiao, and Dai]{li2022bevformer}
Zhiqi Li, Wenhai Wang, Hongyang Li, Enze Xie, Chonghao Sima, Tong Lu, Yu Qiao, and Jifeng Dai.
\newblock {BEVFormer: Learning Bird’s-Eye-View Representation from Multi-Camera Images via Spatiotemporal Transformers}.
\newblock In \emph{Proceedings of the European Conference on Computer Vision (ECCV)}, pages 1--18, 2022.

\bibitem[Liao et~al.(2023)Liao, Chen, Wang, Cheng, Zhang, Liu, and Huang]{MapTR}
Bencheng Liao, Shaoyu Chen, Xinggang Wang, Tianheng Cheng, Qian Zhang, Wenyu Liu, and Chang Huang.
\newblock {MapTR: Structured Modeling and Learning for Online Vectorized HD Map Construction}.
\newblock In \emph{Proceedings of the International Conference on Learning Representations (ICLR)}, pages 1--18, 2023.

\bibitem[Liao et~al.(2024)Liao, Chen, Zhang, Jiang, Zhang, Liu, Huang, and Wang]{maptrv2}
Bencheng Liao, Shaoyu Chen, Yunchi Zhang, Bo Jiang, Qian Zhang, Wenyu Liu, Chang Huang, and Xinggang Wang.
\newblock {MapTRv2: An End-to-End Framework for Online Vectorized HD Map Construction}.
\newblock \emph{International Journal of Computer Vision}, pages 1--17, 2024.

\bibitem[LightwheelAI(2024)]{LightwheelAI}
LightwheelAI.
\newblock Street-gaussians-ns.
\newblock \url{https://github.com/LightwheelAI/street-gaussians-ns}, 2024.

\bibitem[Lilja et~al.(2024)Lilja, Fu, Stenborg, and Hammarstrand]{lilja2024localizationevaluatedataleakage}
Adam Lilja, Junsheng Fu, Erik Stenborg, and Lars Hammarstrand.
\newblock {Localization Is All You Evaluate: Data Leakage in Online Mapping Datasets and How to Fix It}.
\newblock In \emph{Proceedings of the IEEE/CVF Conference on Computer Vision and Pattern Recognition (CVPR)}, pages 22150--22159, 2024.

\bibitem[Liu et~al.(2025)Liu, Huang, Zhang, Yao, Zhang, Wan, Ye, and Zhou]{liu2024ray}
Feng Liu, Tengteng Huang, Qianjing Zhang, Haotian Yao, Chi Zhang, Fang Wan, Qixiang Ye, and Yanzhao Zhou.
\newblock Ray denoising: Depth-aware hard negative sampling for multi-view 3d object detection.
\newblock In \emph{Proceedings of the European Conference on Computer Vision (ECCV)}, 2025.

\bibitem[Luo et~al.(2024)Luo, Weng, Wang, Wu, Li, Weinberger, Wang, and Pavone]{luo2024smerf}
Katie~Z Luo, Xinshuo Weng, Yan Wang, Shuang Wu, Jie Li, Kilian~Q Weinberger, Yue Wang, and Marco Pavone.
\newblock {Augmenting Lane Perception and Topology Understanding with Standard Definition Navigation Maps}.
\newblock In \emph{Proceedings of the IEEE International Conference on Robotics and Automation (ICRA)}, pages 4029--4035, 2024.

\bibitem[Mildenhall et~al.(2020)Mildenhall, Srinivasan, Tancik, Barron, Ramamoorthi, and Ng]{mildenhall2020nerf}
Ben Mildenhall, Pratul~P. Srinivasan, Matthew Tancik, Jonathan~T. Barron, Ravi Ramamoorthi, and Ren Ng.
\newblock Nerf: Representing scenes as neural radiance fields for view synthesis.
\newblock In \emph{ECCV}, 2020.

\bibitem[Philion and Fidler(2020)]{philion2020lift}
Jonah Philion and Sanja Fidler.
\newblock Lift, splat, shoot: Encoding images from arbitrary camera rigs by implicitly unprojecting to 3d.
\newblock In \emph{Computer Vision--ECCV 2020: 16th European Conference, Glasgow, UK, August 23--28, 2020, Proceedings, Part XIV 16}, pages 194--210. Springer, 2020.

\bibitem[Ranganatha et~al.(2024)Ranganatha, Zhang, Venkatramani, Liao, and Christensen]{Ranganatha2024SemVecNet}
Narayanan~Elavathur Ranganatha, Hengyuan Zhang, Shashank Venkatramani, Jing-Yan Liao, and Henrik~I. Christensen.
\newblock Semvecnet: Generalizable vector map generation for arbitrary sensor configurations.
\newblock In \emph{2024 IEEE Intelligent Vehicles Symposium (IV)}, pages 2820--2827, 2024.

\bibitem[Sima et~al.(2023)Sima, Tong, Wang, Chen, Wu, Deng, Gu, Lu, Luo, Lin, and Li]{sima2023_occnet}
Chonghao Sima, Wenwen Tong, Tai Wang, Li Chen, Silei Wu, Hanming Deng, Yi Gu, Lewei Lu, Ping Luo, Dahua Lin, and Hongyang Li.
\newblock Scene as occupancy.
\newblock \emph{Proceedings of the IEEE/CVF International Conference on Computer Vision (ICCV)}, 2023.

\bibitem[Sun et~al.(2020)Sun, Kretzschmar, Dotiwalla, Chouard, Patnaik, Tsui, Guo, Zhou, Chai, Caine, Vasudevan, Han, Ngiam, Zhao, Timofeev, Ettinger, Krivokon, Gao, Joshi, Zhang, Shlens, Chen, and Anguelov]{Sun2020waymo}
Pei Sun, Henrik Kretzschmar, Xerxes Dotiwalla, Aurélien Chouard, Vijaysai Patnaik, Paul Tsui, James Guo, Yin Zhou, Yuning Chai, Benjamin Caine, Vijay Vasudevan, Wei Han, Jiquan Ngiam, Hang Zhao, Aleksei Timofeev, Scott Ettinger, Maxim Krivokon, Amy Gao, Aditya Joshi, Yu Zhang, Jonathon Shlens, Zhifeng Chen, and Dragomir Anguelov.
\newblock {Scalability in Perception for Autonomous Driving: Waymo Open Dataset}.
\newblock In \emph{Proceedings of the IEEE/CVF Conference on Computer Vision and Pattern Recognition (CVPR)}, pages 2443--2451, 2020.

\bibitem[Wang et~al.(2023{\natexlab{a}})Wang, Li, Li, Chen, Sima, Liu, Wang, Jia, Wang, Jiang, Wen, Xu, Luo, Yan, Zhang, and Li]{wang2023openlanev2}
Huijie Wang, Tianyu Li, Yang Li, Li Chen, Chonghao Sima, Zhenbo Liu, Bangjun Wang, Peijin Jia, Yuting Wang, Shengyin Jiang, Feng Wen, Hang Xu, Ping Luo, Junchi Yan, Wei Zhang, and Hongyang Li.
\newblock {OpenLane-V2: A Topology Reasoning Benchmark for Unified 3D HD Mapping}.
\newblock In \emph{Proceedings of the Advances in Neural Information Processing Systems (NeurIPS)}, pages 18873--18884, 2023{\natexlab{a}}.

\bibitem[Wang et~al.(2023{\natexlab{b}})Wang, Zhao, Xu, Chen, Yu, Chang, Yang, and Zhao]{wang2023towards}
Shuo Wang, Xinhai Zhao, Hai-Ming Xu, Zehui Chen, Dameng Yu, Jiahao Chang, Zhen Yang, and Feng Zhao.
\newblock Towards domain generalization for multi-view 3d object detection in bird-eye-view.
\newblock In \emph{Proceedings of the IEEE/CVF Conference on Computer Vision and Pattern Recognition}, pages 13333--13342, 2023{\natexlab{b}}.

\bibitem[Wilson et~al.(2021)Wilson, Qi, Agarwal, Lambert, Singh, Khandelwal, Pan, Kumar, Hartnett, Kaesemodel~Pontes, Ramanan, Carr, and Hays]{Argoverse2}
Benjamin Wilson, William Qi, Tanmay Agarwal, John Lambert, Jagjeet Singh, Siddhesh Khandelwal, Bowen Pan, Ratnesh Kumar, Andrew Hartnett, Jhony Kaesemodel~Pontes, Deva Ramanan, Peter Carr, and James Hays.
\newblock {Argoverse 2: Next Generation Datasets for Self-Driving Perception and Forecasting}.
\newblock In \emph{Proceedings of the Neural Information Processing Systems (NeurIPS) Track on Datasets and Benchmarks}, pages 1--13, 2021.

\bibitem[Wu et~al.(2024)Wu, Chang, Jia, Liu, Wang, and Shen]{wu2024topomlp}
Dongming Wu, Jiahao Chang, Fan Jia, Yingfei Liu, Tiancai Wang, and Jianbing Shen.
\newblock {TopoMLP: A Simple yet Strong Pipeline for Driving Topology Reasoning}.
\newblock In \emph{Proceedings of the International Conference on Learning Representations (ICLR)}, pages 1--12, 2024.

\bibitem[Yan et~al.(2024)Yan, Lin, Zhou, Wang, Sun, Zhan, Lang, Zhou, and Peng]{yan2024streetgs}
Yunzhi Yan, Haotong Lin, Chenxu Zhou, Weijie Wang, Haiyang Sun, Kun Zhan, Xianpeng Lang, Xiaowei Zhou, and Sida Peng.
\newblock Street gaussians: Modeling dynamic urban scenes with gaussian splatting.
\newblock In \emph{ECCV}, 2024.

\bibitem[Zheng et~al.(2023)Zheng, Liu, Wang, and Zhao]{zheng2023cross}
Liangtao Zheng, Yicheng Liu, Yue Wang, and Hang Zhao.
\newblock Cross-dataset sensor alignment: Making visual 3d object detector generalizable.
\newblock In \emph{The Conference on Robot Learning}, 2023.

\bibitem[Zhu et~al.(2021)Zhu, Su, Lu, Li, Wang, and Dai]{zhu2021deformable}
Xizhou Zhu, Weijie Su, Lewei Lu, Bin Li, Xiaogang Wang, and Jifeng Dai.
\newblock {Deformable DETR: Deformable Transformers for End-to-End Object Detection}.
\newblock In \emph{Proceedings of the International Conference on Learning Representations (ICLR)}, pages 1--16, 2021.

\end{thebibliography}
